\pdfoutput=1

\documentclass[11pt]{article}

\usepackage[]{acl}

\usepackage{times}
\usepackage{latexsym}
\usepackage{booktabs}
\usepackage{arydshln}
\usepackage{amsmath}
\usepackage{amsthm,amsmath,amssymb}
\usepackage{multirow}

\usepackage{graphicx} 
\usepackage{stfloats} 
\usepackage{float} 

\usepackage[T1]{fontenc}

\usepackage[utf8]{inputenc}

\usepackage{microtype}

\def\approxprop{%
  \def\p{%
    \setbox0=\vbox{\hbox{$\propto$}}%
    \ht0=0.6ex \box0 }%
  \def\s{%
    \vbox{\hbox{$\sim$}}%
  }%
  \mathrel{\raisebox{0.7ex}{%
      \mbox{$\underset{\s}{\p}$}%
    }}%
}

\DeclareMathAlphabet{\mathbbb}{U}{bbold}{m}{n} 
\newcommand{\ev}{{\mathbb{E}}}
\newcommand{\E}[1]{\ev\left[#1\right]}
\newcommand{\pv}{{\mathbb{P}}}
\newcommand{\Pb}[1]{\pv\left(#1\right)}
\newcommand{\kron}{\mathbbb{1}}

\title{Rethinking and Refining the \textit{Distinct} Metric}

\author{Siyang Liu \and ... \and Yinhe Zheng \\
        Address line \\ ... \\ Address line}

\author{\textbf{Siyang Liu${^{1,2}}$\thanks{\ \ Equal contribution
} , Sahand Sabour$^{1*}$, Yinhe Zheng$^{1,3}$, Pei Ke$^1$, Xiaoyan Zhu$^1$} \\
 \textbf{ Minlie Huang$^1$\thanks{\ \ Corresponding author}} \\
  \small $^1$The CoAI group, DCST, Institute for Artificial Intelligence, State Key Lab of Intelligent Technology and Systems, \\
  \small Beijing National Research Center for Information Science and Technology, Tsinghua University, Beijing 100084, China. \\
  \small $^2$Kuaishou, Beijing, China. \quad \small $^3$ Lingxin AI, Beijing, China. \\
  {\small \tt liusyang641@gmail.com, Sahandfer@gmail.com, zhengyinhe1@163.com} \\ 
  {\small \tt kepei1106@outlook.com, \{zxy-dcs,aihuang\}@tsinghua.edu.cn} \\
}

\begin{document}
\maketitle
\begin{abstract}
Distinct-$n$ score\cite{Li2016} is a widely used automatic metric for evaluating diversity in language generation tasks.
However, we observed that the original approach for calculating distinct scores has evident biases that tend to assign higher penalties to longer sequences.
We refine the calculation of distinct scores by scaling the number of distinct tokens based on their expectations. 
We provide both empirical and theoretical evidence to show that our method effectively removes the biases existing in the original distinct score.
Our experiments show that our proposed metric, \textit{Expectation-Adjusted Distinct (EAD)}, correlates better with human judgment in evaluating response diversity.
To foster future research, we provide an example implementation at   \url{https://github.com/lsy641/Expectation-Adjusted-Distinct}.
\end{abstract}

\section{Introduction}

The diversity of generated texts is an important evaluation aspect for dialogue generation models since most dialogue models tend to produce general and trivial responses (e.g. "I don't know" or "Me too") \cite{Li2016,zhao2017learning}.
Several metrics have been proposed to evaluate the text diversity,
and the \textit{Distinct} score \cite{Li2016} is the most widely applied metric due to its intuitive nature and convenient calculation.
It has become a de facto standard to report the \textit{Distinct} score to compare the performance of different models in terms of response diversity \cite{liu2016not,fan2018hierarchical, sabour2021, wu2021personalized,zhou2021eva,wu2021semantic,zhang2020dialogue,zheng2020pre,wang2020lccc,liu-etal-2021-towards}.
Most previous works follow the initial approach of \citet{Li2016} to calculate the \textit{Distinct} score,
i.e., dividing the number of unique tokens (n-grams) by that of all tokens (n-grams).
However, although reported to be effective, we surprisingly find that this naive approach tends to introduce a higher penalty for longer texts and lead to inaccurate evaluation of text diversity.

\begin{figure}[t]
    \centering
    \includegraphics[width=\linewidth]{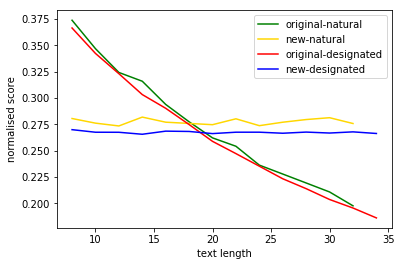}
    \caption{
        \textit{Distinct} (original) and \textit{Expectation-Adjusted Distinct} (new) scores against different sample lengths. In the figure, ``natural'' means that text sets are sampled from a real corpus while ``designated'' means that the sets are sampled from a designated distribution. See details in Section \ref{sec:pre}.
    }
    \label{fig:compare}
\end{figure}

We argue that the scaling factor of \textit{Distinct} requires a comprehensive discussion for two reasons. 
\textbf{First}, prior research in non-computational linguistics has demonstrated the shortcomings of \textit{Distinct}'s scaling approach \cite{malvern2004lexical}. 
We found that early applications of \textit{Distinct} exist in psycholinguistics, where researchers leveraged this metric to assess the language diversity of children with communication disorders \cite{chotlos_1944}. 
Their research showed that as a child speaks more words, \textit{Distinct} experiences an adverse decline since each extra word that the child utters adds to the total number of words, yet it would only increase the number of distinct words if the word had not been used before \cite{malvern2004lexical,chotlos_1944}.
\textbf{Second}, we also discovered an uncommon decline of this metric on both a natural corpus and a designated distribution sampler when the total number of words increases. 
As illustrated in Figure \ref{fig:compare}, the original \textit{Distinct} cannot produce a stable value and experiences a sharp decrease with increasing utterance length in both natural and designated distributions.
However, as a qualified metric needs to support quantitative comparison among different methods, its value should stay invariant when the distribution of the words appearing is determined. 
This result is consistent with the findings of psychologists, indicating an unfair penalty does exist in such a scaling method.

Our contributions are summarized as follows:

\textbf{1.} We investigate the performance of the original \textit{Distinct} and demonstrate that this metric is not sufficiently fair due to its scaling method. 
We also highlight the risks of using this metric for evaluating response diversity.

\textbf{2.} We propose \textit{Expectation-Adjusted Distinct} (\textbf{EAD}), an improved version of \textit{Distinct} based on that the scaling factor should be the expectation of the number of distinct tokens instead. 

\textbf{3.} Human evaluation shows that our metric correlates better with human judgments. We further discuss the drawbacks of this metric and suggest its feasible applications in practice.

\section{Preliminary Discussion about Original Distinct}\label{sec:pre}
To demonstrate the shortcoming of the original \textit{Distinct}, we illustrated the normalised \textit{Distinct} scores on two types of texts at different lengths (Figure \ref{fig:compare}).
The first type of text is sampled from an artificially designated distribution while the other is sampled from a natural language corpus. In detail, we adopted $\Pb{X = k} = \int^{v}_0\frac{\lambda^{k}e^{-\lambda}}{vk!}d\lambda$ as our designated distribution, where $v$ is vocabulary size. In our experiments, we use BERT's vocabulary's size ($v=30522$) \cite{devlin-etal-2019-bert}. In addition, we leveraged  OpenSubtitles\footnote{http://opus.nlpl.eu/OpenSubtitles2018.php} as our natural language corpus. For each length, we sampled 2000 sentences as a set and calculated scores of each set.

As shown in Figure \ref{fig:compare}, We observe that the original \textit{Distinct} scores decrease sharply with increasing utterance length in both distributions. 
We can observe that given the same distribution of words (\textit{original-designated}), lengthier texts will get lower scores than shorter texts. 
We highlighted this problem because it is extremely simple for models to control the length of texts by using decoding tricks, e.g. adjusting the penalty coefficient \cite{vijayakumar2016diverse}. In such cases, it might seem that a model has outperformed other models on this metric. However, as shown by our experiments, it is not reasonable to assume that this model generates more diverse responses.
The same observation can be made for the natural language corpus (\textit{original-designated}). 
As language distribution is more complex than what we are able to formulate, we depicted the performance of the original \textit{Distinct} on 6 famous datasets in \textbf{Appendix}. These cases indicate that the original \textit{Distinct} is not a suitable metric for evaluating diversity.

\section{Improving Original Distinct}\label{sec:improving}

\subsection{Formula Derivation}
The original \textit{Distinct} score \cite{Li2016} is measured as $Distinct = {N}/{C}$, where $N$ is the number of distinct tokens and $C$ is the total number of tokens. To improve the original scaling method, we propose that the scaling factor should be the expectation of the distinct words in the set of generated responses. Hence, the calculation becomes

\begin{equation}
    EAD = \frac{N}{\E{\hat{N}}}.
\end{equation}



Supposing a set of generated responses $R$ with size $S$ to be evaluated, we let $l_{k,i}$ be the $i$\textsuperscript{th} token of $k$\textsuperscript{th} response in $R$ and $t_{k}$ be the length of $k$\textsuperscript{th} response. The expectation  $\mathbf{E}[\hat{N}]$ for $\hat{N}$ distinct words to appear in $R$ would be

\begin{align}
\E{\hat{N}}&=\E{ \sum\limits_{j}^{V} \bigvee\limits_{i,k}^{i=t_k,k=S}\kron_{l_{k,i}=u_j}} \\
&=\sum\limits_{j}^{V} \Pb{\{\bigvee\limits_{i,k}^{i=t_k,k=S}\kron_{l_{k,i}=u_j}\} = 1} \nonumber\\
&=\sum\limits_{j}^{V}(1-\prod_{k}^{S}\Pb{l_{t_k}\neq u_j ,...,l_1\neq u_j })\nonumber,
\end{align}


\noindent where $V$ is the vocabulary size, and $\{u_1,...,u_V\}$ is the set of all tokens in the vocabulary. 

As shown in Equation 2, the calculation requires us to know $\Pb{l_{t_k}\neq u_j ,l_{t_k-1}\neq u_j ,...,l_1\neq u_j}$. Though current models can easily estimate the probability of a word appearing in a sequence, it is hard to calculate the probability of each word that \textbf{never} appears in any position of the sequence. Thus, there is no efficient way to calculate $\Pb{l_{k,t}\neq u_j ,...,l_{k,1}\neq u_j )}$. In addition, different language distributions have different $\pv$, which leads to different expectations and make the metric less general. Thus, we measure the upper bound of response diversity (i.e. a set of generated responses where each token appears with equal probability) to calculate this expectation. We hypothesize that the scaling effect of the upper bound is approximately proportional to that of other sets of generated responses; therefore, it can replace the original scaling factor. 

As mentioned above, we hypothesize 
\begin{equation}
\E{\hat{N}} \approxprop \E{\hat{N_{upper}}}\nonumber,
\end{equation}

\noindent where $\E{\hat{N_{upper}}}$ can be calculated as


\begin{align}
\E{\hat{N_{upper}}} &= \sum\limits_{j}^{V}(1-\prod_{k}^{S}\prod_{i}^{t_k}\Pb{l_{k,i}\neq u_j })\nonumber\\
&= V[1-(\frac{V-1}{V})^C].
\end{align}


Thus, the \textit{EAD} score is calculated as: 

\begin{equation}
    EAD = \frac{N}{V[1-(\frac{V-1}{V})^C]}.
\end{equation}
We discuss more details on the formula's properties and the vocabulary size in the \textbf{Appendix}.

\subsection{Experimental Verification}

\subsubsection{Evaluation Approach}
We collect responses from ten dialogue generation methods as reported by \citet{adaLab}, and compare \textit{EAD} with the original uni-gram \textit{Distinct} \cite{Li2016}. More details of these ten methods can be find in Appendix.

We follow previous works \cite{tao2018ruber, sellam-etal-2020-bleurt} to evaluate the correlation of each automatic metric with human judgments. Specifically, the Pearson, Spearman, and Kendall’s Tau correlation coefficients are reported. Pearson’s correlation estimates linear correlation while Spearman’s and Kendall’s correlations estimate monotonic correlation, with Kendall's correlation being usually more insensitive to abnormal values. We used SciPy\footnote{https://docs.scipy.org/doc/scipy/reference/stats.html} for correlation calculation and significance test

\subsubsection{Datasets}
Our experiments use two open-domain dialog generation benchmark datasets: DailyDialog\cite{li-etal-2017-dailydialog}, a high-quality dialog dataset collected from daily conversations, and OpenSubtitles\footnote{http://opus.nlpl.eu/OpenSubtitles2018.php}, which contains dialogs collected from movie subtitles (see Table \ref{table:dataset} for more details). We follow the data processing procedures reported by \citet{adaLab}.

\begin{table}[htbp]
  \centering
  \scalebox{0.8}{
    \begin{tabular}{llll}\toprule
          & Train & Val   & Test \\\midrule
    DailyDialog & 65.8K & 6.13K & 5.80K \\
    OpenSubtitles & 1.14M & 20.0K & 10.0K \\\bottomrule
    \end{tabular}%
    }
  \caption{Dataset Statistics}
  \label{table:dataset}%
\end{table}%

\subsubsection{Preliminary Observations}

Based on the obtained results (check Table \ref{table:Results}), it can be observed that \textit{Expectation-Adjusted Distinct} has a clear edge over the original \textit{Distinct}: \textbf{first}, the contrast between diversity of generated responses for different methods is highlighted more effectively by \textit{EAD} (e.g. though AdaLab gets the highest diversity score using \textit{Distinct} (3.96), its difference from other methods is not as evident as its \textit{EAD} score (9.63)); \textbf{second}, contrary to Distinct, \textit{EAD} provides a more accurate evaluation of response diversity. For instance, the Distinct scores for CP and UL are both 2.35 while responses generated by UL are found to be more diverse than CP using \textit{EAD} (5.35 > 5.08). Given that the average length of responses generated by FL is larger than CP, \textit{Distinct}'s bias towards models that generate shorter sentences becomes evident. These observations are consistent for both datasets.

\defcitealias{szegedy2016rethinking}{LS(2016)}
\defcitealias{lin2017focal}{FL(2017)}
\defcitealias{jiang2019improving}{Face(2019)}
\defcitealias{choi-etal-2020-f}{F\textsuperscript{2}(2020)}
\defcitealias{pereyra2017regularizing}{CP(2017)}
\defcitealias{welleck2019neural}{UL(2019)}
\defcitealias{he-glass-2020-negative}{NL(2020)}
\defcitealias{li2019data}{D2GPo(2019)}
\defcitealias{chen-etal-2020-distilling}{CE(2020)}
\defcitealias{adaLab}{AdaLab(2021)}
\begin{table*}[!t]
  \centering
  \resizebox{0.8\textwidth}{!}{
    \begin{tabular}{l c  c c c c c c c}
        \toprule
        \multirow{2}{*}{\textbf{Method}} &
        \multicolumn{4}{c}{\textbf{DailyDialog}} &  
        \multicolumn{4}{c}{\textbf{OpenSubtitles}} \\\cmidrule(lr){2-5} \cmidrule(lr){6-9}
        & Avg Length  & \textit{Distinct}  & \textit{EAD}  & Human & Avg Length & \textit{Distinct}  & \textit{EAD} & Human\\ \midrule

        \citetalias{lin2017focal} & 9.33 & 2.38  & 5.09 &5.18 & 8.56  & 3.19 & 9.51 &4.91\\ 
        \citetalias{he-glass-2020-negative} & 9.99 & 1.66 & 3.70 &4.54 & 8.40 & 3.24 & 9.52 &5.02 \\ 
        \citetalias{pereyra2017regularizing} & 8.67  & 2.35 & 4.80 &5.08 & 8.74  & 3.11 & 9.44 &5.20\\
        \citetalias{szegedy2016rethinking} & 8.50 & 1.48 & 2.98 &5.28 & 9.04  & 2.77 & 8.64 & 5.04\\ 
        \citetalias{li2019data} & 9.15  & 1.26 & 2.65 &4.92 & 8.77 & 2.07 & 6.32 &4.89 \\
        \citetalias{chen-etal-2020-distilling} & 8.29 & 1.67 & 3.31 &4.14 &  9.21  & 2.55 & 8.08 &4.95\\
        \citetalias{choi-etal-2020-f} & 8.71 & 1.40 & 2.87 &4.88 & 8.60 & 2.89 & 8.67 &4.52\\
        \citetalias{welleck2019neural} & 9.93 & 2.35 & 5.23 &5.35 & 8.09  & 2.84 & 8.10 &5.00\\ 
        \citetalias{jiang2019improving} & 10.62 & 1.63 & 3.79 &5.26 & 9.11 & 3.31 & 10.41 &5.31\\ 
        \citetalias{adaLab} & 11.30 & 3.96 & 9.63 &5.92 & 8.12  & 4.78 & 13.68 &5.32\\ \midrule
        \textbf{Pearson} & - &\textbf{0.67}\ddag &\textbf{0.70}\ddag &\textbf{1.00} &- &\textbf{0.56}\dag &\textbf{0.60}\dag &\textbf{1.00} \\
        \textbf{Spearman} & - &\textbf{0.42}\dag &\textbf{0.62}\dag &\textbf{1.00} &- &\textbf{0.62}\dag &\textbf{0.65}\ddag &\textbf{1.00} \\
        \textbf{Kendall’s Tau} & - &\textbf{0.27} &\textbf{0.47}\dag &\textbf{1.00} &- &\textbf{0.51}\ddag &\textbf{0.56}\ddag &\textbf{1.00} \\ 

        \bottomrule
    \end{tabular}}

  \caption{
    Results of automatic and human evaluation on corpus-level diversity methods. Pearson/Spearman/Kendall’s Tau indicates the   Pearson/Spearman/Kendall’s Tau correlation respectively. The correlation scores marked with \dag  (i.e., $p$-value$<0.1$) and \ddag  (i.e., $p$-value$<0.05$) indicate the result significantly correlates with human judgments.
  }
  \label{table:Results}
\end{table*}

\subsubsection{Correlation Results} 
We recruited crowdsourcing workers to evaluate the diversity of the selected methods\footnote{See Appendix for more details on the human evaluation interface}. For each method, we randomly sampled 100 subsets of 15 responses from their set of generated responses. Response sets of all methods, given the same query set, were packaged together as an evaluation set. We asked each crowdsourcing worker to assign a diversity score to every response group in the evaluation set. Each group was evaluated by at least 3 workers. For ensuring the quality of our annotations, we calculated the score of each set as the average of workers' scores and filtered out workers whose scores had an insufficient correlation with the average (Pearson Correlation < 0.65). We acknowledge that building a scoring standard for annotating language diversity is challenging. Hence, we did not require our workers to give an absolute score for each set. Instead, we asked them to highlight the contrast between different sets by scoring values that linearly reflect the response diversity difference between the sets.  For instance, the two sets of scores $\{1,2,2\}$ and $\{2,5,5\}$ show the same evaluation since the same contrast is shown. We then normalized the scores to the [0-10] range. 

Then, we calculated the correlation between the Distinct scores with the crowdsourced values for all the methods. The results  are provided in Table \ref{table:Results}. The evaluation results indicate that our proposed \textit{EAD} is more consistent with human judgments for measuring response diversity, as it shows the highest correlation with human evaluations among all correlation metrics (Pearson/ Spearson/ Kendall’s Tau) on both datasets.

\section{EAD in Practice}
As \textit{EAD} is based on the idealized assumption that does not take language distribution into account, we further discuss this problem and propose a potential practical way of \textit{Expectation-Adjusted Distinct} in real situations. 
Before applying EAD, it is necessary to explore the relationship between score and text length (Figure \ref{fig:compare}) and check the performance of \textit{EAD} on the training data. To our knowledge, if the training data is from large-scale open-domain sources such as OpenSubtitles and Reddit, \textit{EAD} can maintain its value on different lengths. Hence, it can be directly used for evaluating models trained on these datasets. However, we found our experiments on datasets such as Twitter showed a decline in \textit{EAD} on lengthier texts. This is probably because input length limitations on these platforms (e.g. 280 words on Twitter), which induces users to say as much information as possible within a shorter length. In these situations, it is unfair to use \textit{EAD} to evaluate methods that tend to generate lengthier texts.

\section{Related Work}

\citet{Li2016} proposed \textit{Distinct}, calculated as the number of distinct tokens divided by the total number of tokens. 
This automatic metric is designed to evaluate the diversity of texts, and it has been widely used in developing various text generation tasks, such as dialogue generation \cite{wu2021transferable,zheng2021mmchat,zheng2021stylized,zheng2019personalized} or story generation \cite{guan2021openmeva}.
However, as we showed in Figure \ref{fig:compare}, it is an unfair indicator as it is affected by the sample length. This causes a bias against models which tend to generate longer sentences.

There exist other metrics for evaluating diversity but none are as widely-used as \textit{Distinct} \cite{Zhu2018, Xu2018}. Specifically, Self-BLEU proposed by \citet{Zhu2018} is extremely time-consuming as its computation complexity is $O(n^2)$, where $n$ denoted the size of the test set.

\section{Conclusion}
In this paper, we present an improved variation of the Distinct metric, which is a widely-used measure for evaluating response diversity in dialog systems. We provide the theoretical formulation and empirical evaluation of our proposed metric (\textit{Expectation-Adjusted Distinct}). The results demonstrated that \textit{Expectation-Adjusted Distinct} has a higher correlation with human evaluation in comparison with other metrics. The proposed metric is not limited to dialogue generation models but also suitable to evaluate text generation tasks where diversity matters.

\section{Acknowledgements}
This work was supported by the National Science
Foundation for Distinguished Young Scholars (with
No. 62125604) and the NSFC projects (Key project
with No. 61936010 and regular project with No.
61876096). This work was also supported by the
Guoqiang Institute of Tsinghua University, with
Grant No. 2019GQG1 and 2020GQG0005.
We were grateful to Dr. Xiangxiang Xu at MIT for his help in mathematical formulation.


\bibliography{anthology,custom}
\bibliographystyle{acl_natbib}

\appendix

\section{Comparison on More Datasets}
To demonstrate the shortcomings of the original Distint metric, we illustrate original Distinct on 6 datasets: Persona-chat \cite{zhang2018personalizing}, Ubuntu Dialog Corpus \cite{lowe2015ubuntu}, DailyDialog, Topic-Chat \cite{gopalakrishnan2019topical}, Empathetic Dialogs \cite{rashkin2018towards}, Wizard of Wikipedia \cite{dinan2018wizard}, Reddit \cite{serban2015survey}, and Twitter \cite{ritter2010unsupervised} (Figure \ref{fig:compare}). It can be observed that with an increasing sample length, the original Distinct score tends to follow a linear decline while the proposed metric maintains its consistency.

 \begin{figure*}[h]
    \centering
    \includegraphics[width=\linewidth]{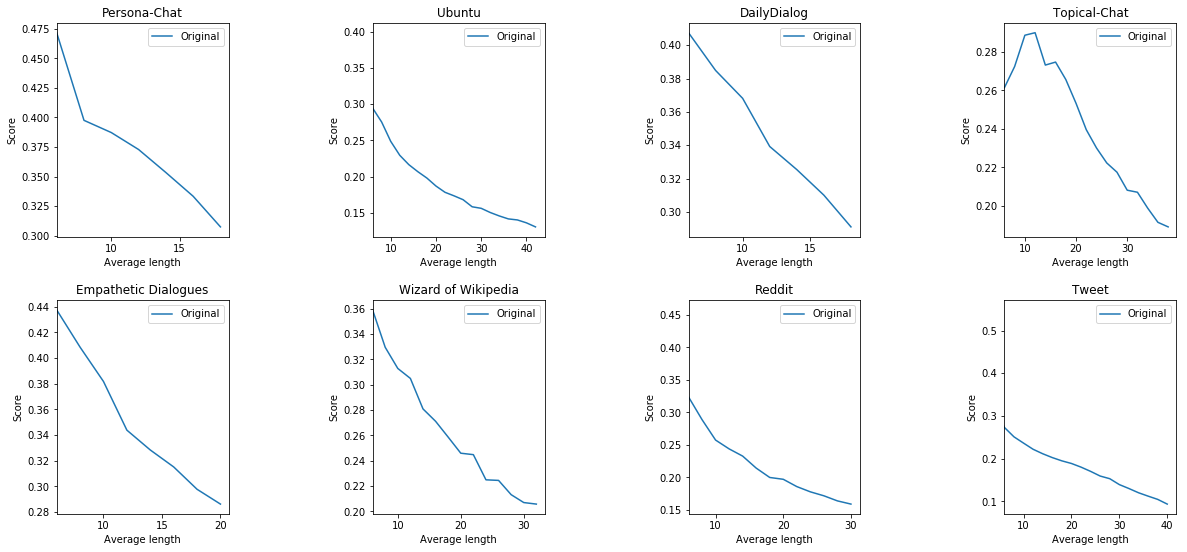}
    \caption{
        Original scores against different sample lengths. The dotted lines are the actual curves for each score while the lines are slope-intercept graphs of the curves. Each score is calculated based on 10 sets of 2000 randomly sampled responses with the same certain length. 
    }
    \label{fig:morecompare}
\end{figure*}

\section{Property Discussion}





\noindent\textbf{Formula Property 1.} \textit{Expectation-Adjusted Distinct} increases faster as $C$ is increasing, but its incremental rate converges to $\frac{1}{V}$, as shown by its derivative below:

\begin{align}
    \frac{\mathrm{d} EAD}{\mathrm{d} N} = \frac{1}{V[1-(\frac{V-1}{V})^C]}\\
    \lim\limits_{C\to+\infty} \frac{\mathrm{d} EAD}{\mathrm{d} N}  = \frac{1}{V}
\end{align}

\noindent whereas in the original Distinct, we have

\begin{align}
    \frac{\mathrm{d} Distinct }{\mathrm{d} N} = \frac{1}{C}
\end{align}
\\
We can see from the original metric that the bigger $C$ is, the slower the original Distinct increases. It is the reason why this metric is not fair to those models that tend to generate longer sentences. 

\noindent\textbf{Formula Property 2.} \textit{Expectation-Adjusted Distinct} converges to $\frac{N}{V}$ ($\leq 1$) as $C$ increases. 

\begin{align}
    \lim\limits_{C\to+\infty}{EAD} &=\lim\limits_{C\to+\infty}\frac{N}{V[1-(\frac{V-1}{V})^C]}\\
    &= \frac{N}{V} <= 1,
\end{align}

\noindent where $\frac{N}{V[1-(\frac{V-1}{V})^C]} \in [0,+\infty]$. 
Theoretically, \textit{Expectation-Adjusted Distinct} can have values larger than 1 (e.g. when $N=V$), which is an extremely rare case in practice: as we utilized the upper bound for measuring the expectation, it is exceptionally hard for $N$ to obtain an equal value to or an even greater value than $\mathbf{E}(\hat{N_{upper}})$.

\section{Details of Human Evaluation}
\begin{figure*}[hpb]
    \centering
    \includegraphics[width=1\linewidth]{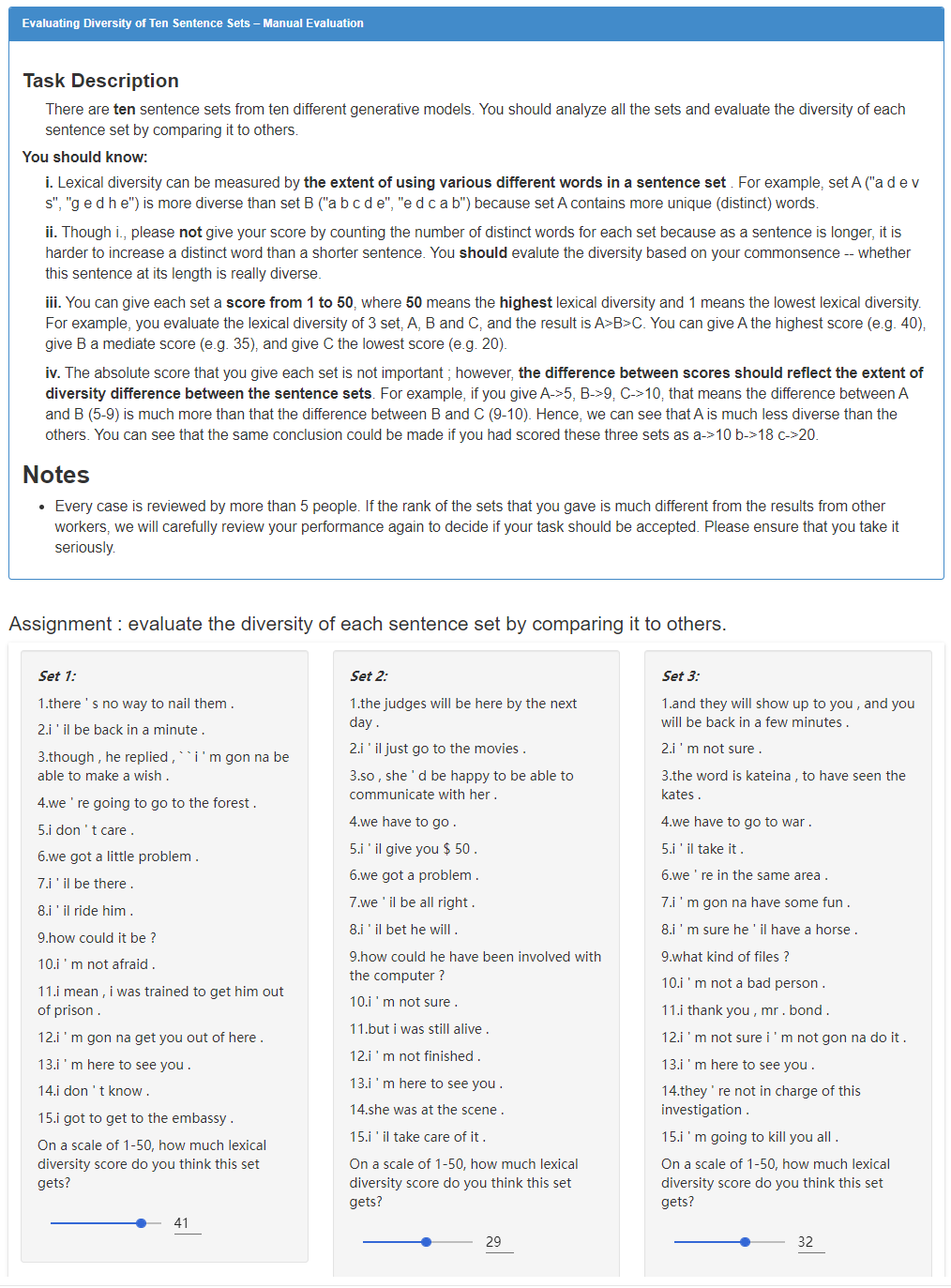}
    \caption{
        Interface of Human Evaluation
    }
    \label{fig:humanInterface}
\end{figure*}

Our created human evaluation interface is provided in Figure \ref{fig:humanInterface}.

\section{How to Determine Vocabulary Size}
As we discussed the properties of \textit{Expectation-Adjusted Distinct}, vocabulary size makes little impact on changing its value when it has reached a large number (usually more than 30000), so it is not necessary to measure an exact value. To compare different methods, it is recommended to use a common vocabulary size, (such as BERT's 30522) \cite{devlin-etal-2019-bert}. It is also reasonable to calculate the vocabulary size of a dataset by NLTK tokenizer, when research focuses on a specific dataset. For non-english corpora, we recommend researchers to determine a vocabulary size following \citet{xu-etal-2021-vocabulary}.

\section{Details of Evaluated Methods}
\citet{adaLab} proposed a novel adaptive label smoothing method for diversified response generation. Their experiments were conducted on the DailyDialog and OpenSubtitles datasets, using 9 recent methods for diverse response generation as their baselines (similar to what we demonstrated in our paper). \citet{adaLab} used a transformer-based sequence-to-sequence model \cite{vaswani2017attention} as the backbone of their model, and most of their hyper-parameters follow \citep{cai-etal-2020-data}. In addition, both the encoder and the decoder contain 6 transformer layers with 8 attention heads, and the hidden size is set to 512. BERT's WordPiece tokenizer \cite{devlin-etal-2019-bert} and Adam optimizer \cite{kingma2015adam} are used for training their models with random initialization and a learning rate of 1e-4.

\end{document}